\documentclass[10pt,twocolumn,letterpaper]{article}

\usepackage{cvpr}
\usepackage{times}
\usepackage{epsfig}
\usepackage{graphicx}
\usepackage{amsmath}
\usepackage{amssymb}
\usepackage{booktabs}
\usepackage{multirow}
\usepackage{gensymb}
\usepackage{xspace}
\usepackage[pagebackref=true,breaklinks=true,letterpaper=true,colorlinks,bookmarks=false]{hyperref}

\usepackage{xcolor}
\definecolor{ForestGreen}{rgb}{0.0, 0.5, 0.0}
\definecolor{DarkRed}{rgb}{0.7, 0.0, 0.0}
\definecolor{DarkMagenta}{rgb}{0.55, 0.0, 0.55}

\newcommand{\paul}[1]{{\color{DarkMagenta}#1}}
\newcommand{\vincentrmk}[1]{{\color{ForestGreen}\bf #1}}
\newcommand{\stefanrmk}[1]{{\color{DarkRed}\bf #1}}
\newcommand{\paulrmk}[1]{{\color{DarkMagenta}\bf #1}}
\newcommand{\comment}[1]{}

\newcommand{\red}[1]{\textcolor{red}{#1}}
\newcommand{\blue}[1]{\textcolor{blue}{#1}}
\newcommand{\imgspace}[0]{\hspace{-0.3cm}}
\newcommand{\asus}[0]{AsusXtionPROLive\xspace}
\newcommand{\ptgrey}[0]{PtGreyBlackfly\xspace}

\cvprfinalcopy 

\def\cvprPaperID{3380} 

\ifcvprfinal\fi
\begin{document}

\title{On Pre-Trained Image Features and Synthetic Images for Deep Learning}

\author{Stefan Hinterstoisser\\
X\\
{\tt\small hinterst@google.com}
\and
Vincent Lepetit\\
Universit\'e de Bordeaux\\
{\tt\small vincent.lepetit@u-bordeaux.fr}
\and
Paul Wohlhart\\
X\\
{\tt\small wohlhart@google.com}
\and
Kurt Konolige\\
X\\
{\tt\small konolige@google.com}
}

\maketitle

\begin{abstract}
   Deep Learning  methods usually require huge amounts  of training data
   to  perform at  their  full  potential, and  often  require expensive  manual
   labeling.   Using synthetic  images  is therefore  very  attractive to  train
   object detectors,  as the  labeling comes for  free, and  several approaches
   have been proposed to combine  synthetic and real images for training.

   In this  paper, we show that  a simple trick is  sufficient to train
     very effectively  modern object  detectors with  synthetic images  only: We
     'freeze' the  layers responsible for  feature extraction to  generic layers
     pre-trained on real images, and train  only the remaining layers with plain
     OpenGL rendering. Our  experiments with very recent  deep architectures for
     object   recognition~(Faster-RCNN,    R-FCN, Mask-RCNN)   and   image   feature
     extractors~(InceptionResnet and Resnet) show this simple approach performs
     surprisingly well.

\comment{
  Deep  Learning approaches  are extremely  data hungry  and usually
    require expensive  manual labeling.  Missing labeled  data is very
    often one of the biggest hurdles to be successful and therefore it
    would be  of great benefit if  we could train our  approaches with
    synthetic  data  since  in  that case  labeling  comes  for  free.
    Unfortunately, it is still extremely difficult to render synthetic
    data with  the same  domain specific properties  as real  data and
    thus, the  transfer from  synthetically trained approaches  to the
    real world  usually results in  low performance. \newline  In this
    paper, we show a simple yet  very effective trick to overcome this
    problem  for  object  detection.   In detail,  we  use  an  object
    detection net  where the feature  extractor is pre-trained  on real
    classification data,  freeze the feature extractor  and only allow
    the  remaining layers  of  the  net to  adapt  its weights  during
    learning.   Using  this  trick  allows  us  to  use  plain  OpenGL
    rendering  and  to  avoid efforts  in  generating  photo-realistic
    rendering  and scene  composing.  Our  method allows  to train  on
    purely synthetic generated data and  we show that our trick boosts
    the performance significantly.
   }
\end{abstract}

\section{Introduction}
\label{sec:introduction}

\begin{figure}[ht]
\begin{center}
\begin{tabular}{cc}
\includegraphics[width=0.5\linewidth]{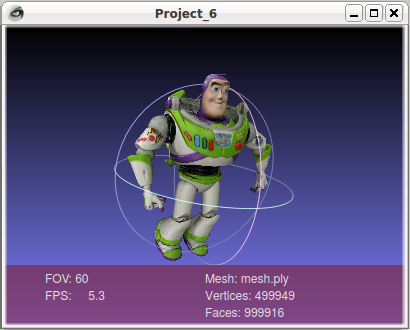} & \imgspace
\includegraphics[width=0.5\linewidth]{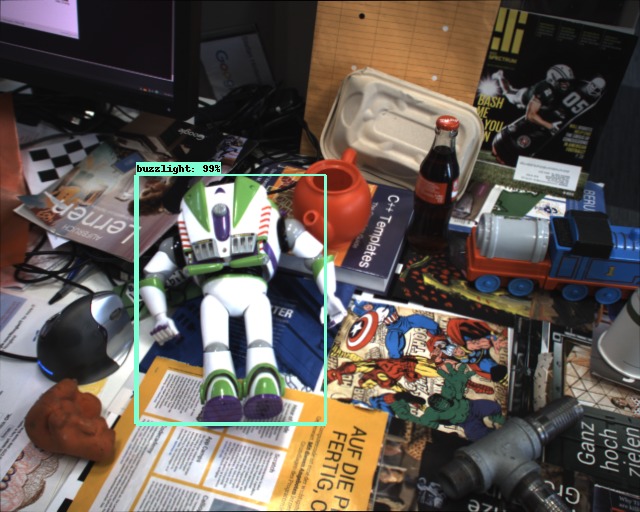} \\
\includegraphics[width=0.5\linewidth]{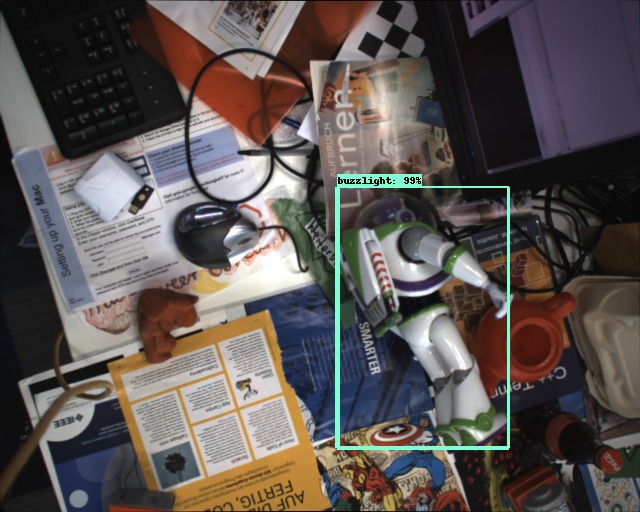} & \imgspace
\includegraphics[width=0.5\linewidth]{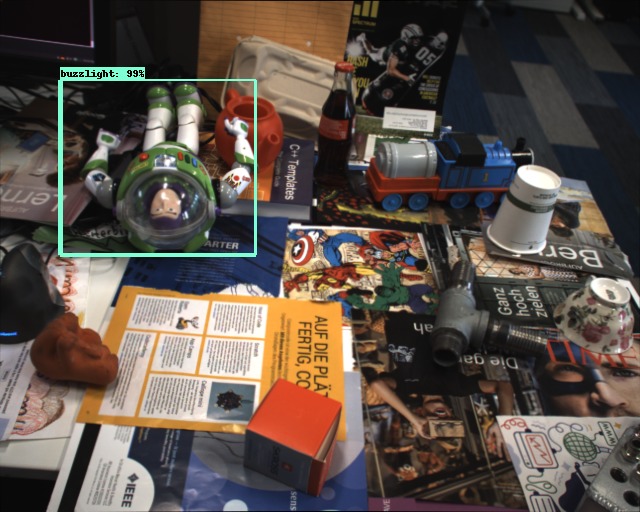} \\
\end{tabular}
\end{center}
\caption{\label{fig:overview}
We show that feature extractor layers from modern object detectors pre-trained on real images 
can be used on synthetic images to learn to detect objects
in real images. The top-left image shows the CAD model we used to
learn to detect the object in the three other images. }
\comment{
In this  paper,  we show  how a  simple
  trick enables  learning deep object detectors  purely from synthetic
  data.  In the  upper left image we  see one of the 3D  CAD models we
  learn from and  the remaining images show the  object detected under
  various  poses in  a heavily  cluttered environments.  We achieve  a
  detection performance  up to  $95\%$ of a  detector trained  on real
  data.}
\end{figure}

\begin{figure*}[ht]
\begin{center}
\begin{tabular}{cc}
\includegraphics[width=0.4\linewidth]{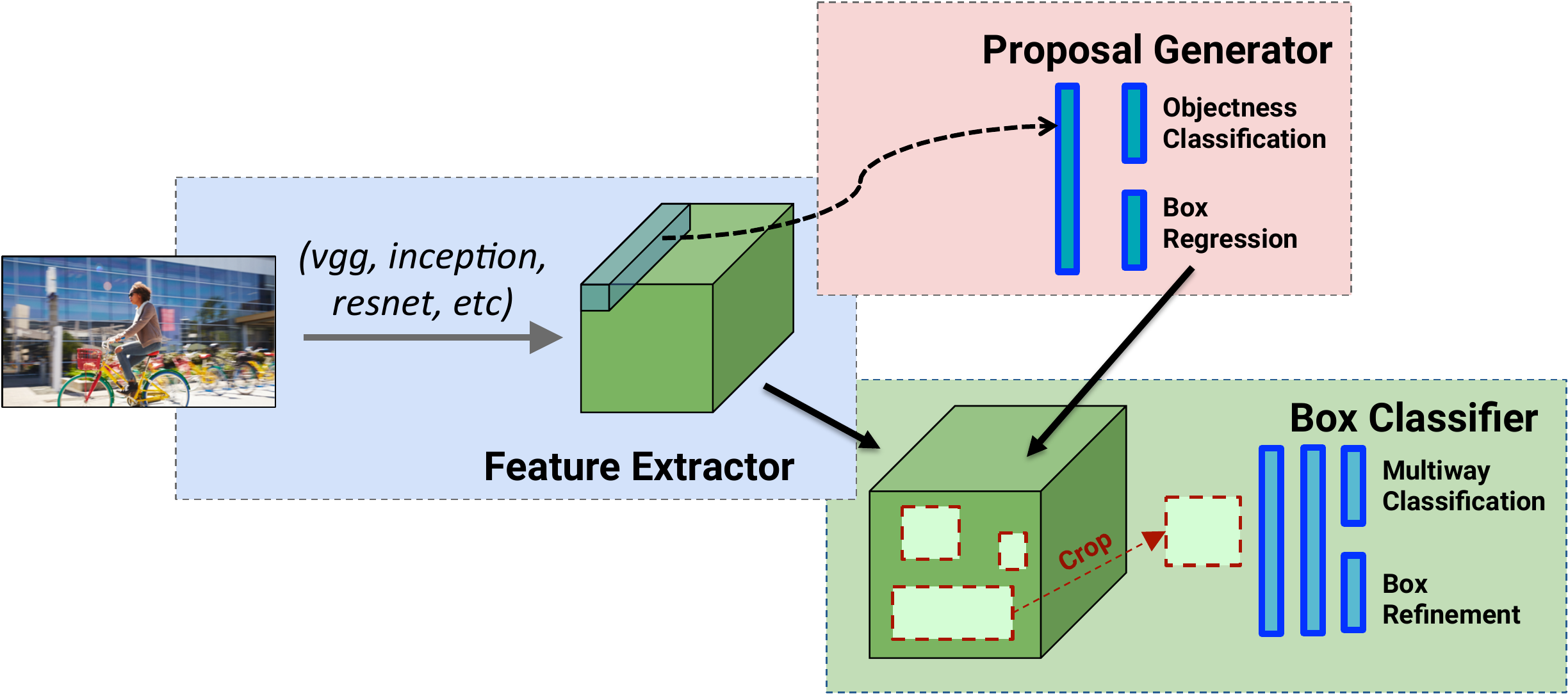} &
\includegraphics[width=0.4\linewidth]{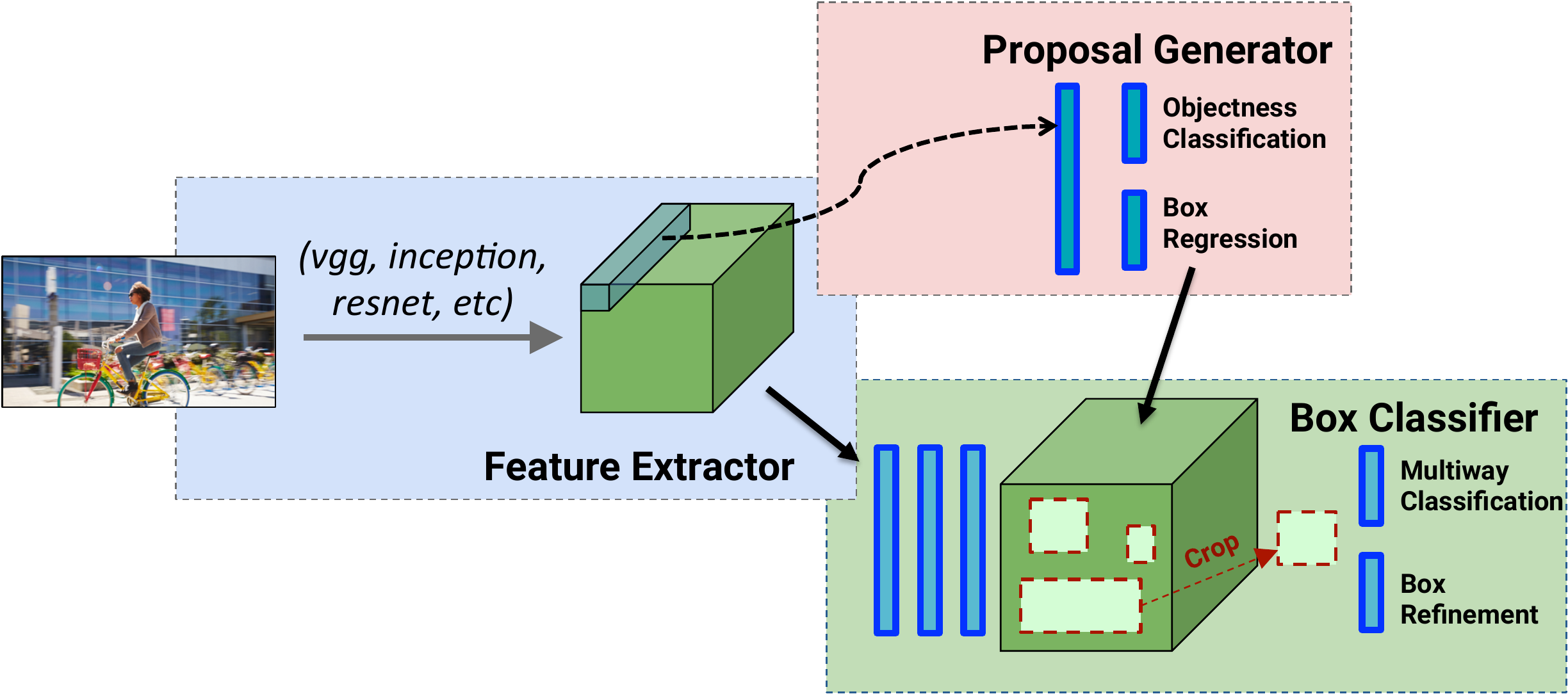} \\
Faster-RCNN &
R-FCN \\
\end{tabular}
\end{center}
\caption{\label{fig:metaarchitecture} The architectures of two recent object
  detectors with their feature extractors isolated as described
  in~\cite{Huang17} (Figure taken from \cite{Huang17}).}
\end{figure*}

The  capability  of detecting  objects  in  challenging  environments is  a  key
component for  many computer vision  and robotics task.  Current  leading object
detectors---Faster-RCNNs~\cite{faster_rcnn},  SSD~\cite{ssd},  RFCN~\cite{rfcn},
Yolo9000~\cite{redmon16}---all rely  on convolutional neural  networks.  However,
to perform  at their best, they  require huge amounts of  labeled training data,
which is usually time consuming and expensive to create.

Using synthetic images  is therefore very attractive to  train object detectors,
as the  labeling comes for  free.  Unfortunately, synthetic  rendering pipelines
are  usually unable  to reproduce  the statistics  produced by  their real-world
counterparts.  This is  often referred to as the 'domain  gap' between synthetic
and  real  data  and  the  transfer  from one  to  another  usually  results  in
deteriorated performance, as observed in \cite{Tobin17} for example.

Several  approaches have  tried  to  overcome this  domain  gap.  For  instance,
\cite{Dwibedi17,Georgakis17,Rad17c}  use synthetic  images in  addition to  real
ones to boost  performance.  While this usually results in  good performance, it
is still dependent on real world labeled data.  Transfer learning approaches are
also possible~\cite{Rozantsev17,bousmalis2016domain,ganin2016domain}, however  they also require real  images of the
objects to  detect.  \cite{Gupta16,  Alhaija17} create  photo-realistic graphics
renderings and \cite{Gupta16, Alhaija17, Georgakis17, Varol17} compose realistic
scenes which both shows to improve performance.  Unfortunately, these strategies
are usually difficult to engineer, need domain specific expertise and require some
additional data  such as illumination  information and scene labeling  to create
realistic scenes.  \cite{Tobin17} uses 'domain randomization' to narrow the gap.
While this has shown very promising  results, it has mainly been demonstrated to
work       with      simple       objects       and      scenarios.        Other
works~\cite{Shrivastava16,Bousmalis17}      use      Generative      Adversarial
Networks~(GANs) to  remove the domain gap,  however, GANs are still very brittle
and hard to train, and to our knowledge they have not been used for detection
tasks yet.




 

In  this   paper  we   consider  a  simple   alternative  solution.    As  shown
by~\cite{Huang17}  and illustrated  in Fig.~\ref{fig:metaarchitecture},  many of
today's modern feature extractors can be split into a feature extractor and some
remaining  layers that  depend on  the meta-architecture  of the  detector.  Our
claim is twofold: a) the pre-trained  feature extractors are already rich enough
and do not need to be retrained  when considering new objects to detect; b) when
applied to an  image synthetically generated using  simple rendering techniques,
the feature  extractors work as a  ``projector'' and output image  features that
are close to real image features.

Therefore, by freezing the weights of feature extractor pre-trained on real data
and by only adapting the weights of the remaining layers during training, we are
able to train state-of-the-art object detectors purely on synthetic data.  While
using  pre-trained  layers for  feature  extraction  is  not new  (for  example,
VGG~\cite{vgg}   has  been   used  extensively   for  this   purpose),  our
contribution is therefore  to show that this approach is  also advantageous when
training with synthetic data.  We also  show in this paper that this observation
is fairly general and we give  both qualitative and quantitative experiments for
different   detectors---Faster-RCNN~\cite{faster_rcnn},   RFCN~\cite{rfcn}   and
Mask-RCNN~\cite{mask_rcnn}---and       different        feature       extraction
networks---InceptionResnet~\cite{inception_resnet}                           and
Resnet101~\cite{resnet101}.

Furthermore, we show that different cameras have different image statistics that
allow different  levels of performance  when re-trained on  synthetic data.
We will demonstrate that performance  is significantly boosted for these cameras
if our simple approach is applied.

In the  remainder of the  paper we  first discuss  related work, describe  how we
generate synthetic data,  demonstrate the domain gap between  synthetic and real
data, and show how to boost  object detection by freezing the pre-trained feature
extractor  during  training.  Finally,  we  show  qualitative  and  quantitative
experiments.

\section{Related Work}
\label{sec:related_work}

Mixing real  and synthetic data to improve detection  performance is a
well      established      process.       Many      approaches      such      as
\cite{Dwibedi17,Georgakis17,Su15}, to mention only  very recent ones, have shown
the usefulness of  adding synthetic data when real data  is limited. In contrast
to \cite{Dwibedi17,Georgakis17} which use real masked image patches, \cite{Su15}
uses 3D CAD  models and a structure-preserving deformation  pipeline to generate
new  synthetic models  to prevent  overfitting.  However,  while these
approaches  obtain better  results compared  to detectors  trained on  real data
only, they still require real data.


In order to avoid expensive labeling in terms of time and money, some approaches
learn object detectors purely from synthetic data.  For instance, a whole line of
work uses  photo-realistic rendering~\cite{Alhaija17,Gupta16} and  complex scene
composition~\cite{Gupta16, Alhaija17, Georgakis17, Varol17}  to achieve
good   results,   and   \cite{MovshovitzAttias16}    stresses   the   need   for
photo-realistic rendering.  Some  approaches even use physics  engines to enable
realistic  placing   of  objects~\cite{Mitash17}.   This   requires  significant
resources and highly elaborate pipelines that are difficult to engineer and need
domain  specific  expertise~\cite{Richter_2016_ECCV}.   Furthermore,  additional
effort  is   needed  to   collect  environment  information   like  illumination
information~\cite{Alhaija17} to produce photo-realistic  scenes.  For real scene
composition,  one also  needs to  parse real  background images  semantically in
order to place the objects meaningful into the scene.

This usually  needs manual post-processing  or labeling which is  both expensive
and time  consuming.  While  these graphics  based rendering  approaches already
show some of the advantages of learning from synthetic data, they usually suffer
from the domain gap between real and synthetic data.

To address  this, a  new line of  work~\cite{Dwibedi17,Georgakis17,Rad17c} moves
away from  graphics based renderings  to composing real images.   The underlying
theme is to paste masked patches of  objects into real images, and thus reducing
the dependence on graphics renderings.  This approach has the advantage that the
images  of the  objects are  already in  the right  domain---the domain  of real
images---and thus, the domain gap between  image compositions and real images is
smaller than  the one of graphics  based rendering and real  images.  While this
has shown  quite some  success, the amount  of data is  still restricted  to the
number of images taken from the object  in the data gathering step and therefore
does not allow to come up with new  views of the object.  Furthermore, it is not
possible to generate new illumination  settings or proper occlusions since shape
and depth are usually not available.  In addition, this approach is dependent on
segmenting out  the object from  the background  which is prone  to segmentation
errors when generating the object masks.

Recently, several approaches~\cite{Shrivastava16,Bousmalis17}  tried to overcome
the domain gap  between real and synthetic data by  using generative adversarial
networks~(GANs).  This way they produced  better results than training with real
data.  However, GANs  are hard to train  and up to now, they  have mainly shown
their usefulness on regression tasks and not on detection applications.

Yet      another       approach      is      to      rely       on      transfer
learning~\cite{Rozantsev17,bousmalis2016domain,ganin2016domain},  to  exploit  a
large amount of available data in a  source domain, here the domain of synthetic
images, to correctly  classify data from  the target domain,  here the
domain of real images, for which the amount of training data is limited. This is
typically done  by tighting two predictors  together, one trained on  the source
domain, the other on the target domain or by training a
  single predictor on the two domains. This is a general approach as the source
and target domains can be very different, compared to synthetic and real images,
which are more related to each other. In this paper, we exploit this relation by applying
the same feature extractor to the two domains. However, in contrast to~\cite{Rozantsev17,bousmalis2016domain,ganin2016domain} we do not need any real
images of the objects of interest in our approach.

\section{Method}

In this section,  we will present our simple synthetic  data generation pipeline
and describe how we change  existing state-of-the-art object detectors to enable
them to  learn from synthetic  data.  In this context,  we will focus  on object
instance   detection.   Throughout   this  paper,   we  will   mainly  rely   on
Faster-RCNN~\cite{faster_rcnn}   since  it   demonstrated  the   best  detection
performance among a whole family of object detectors as shown in~\cite{Huang17}.
However, in order to show the generability of our approach, we will also present
additional   quantitative   and   qualitative   results   of   other   detectors
(RFCN~\cite{rfcn}          and           Mask-RCNN~\cite{mask_rcnn})          in
Section~\ref{sec:other_detectors}.

\comment{
\vincentrmk{we need to mention https://arxiv.org/pdf/1411.7911.pdf somewhere}
}

\subsection{Synthetic Data Generation Pipeline}
\label{lab:synthetic_data_generation}

\begin{figure*}[ht]
\begin{center}
\includegraphics[trim={0.0cm 5.7cm 0.0cm 0.0cm},clip=true,width=1\linewidth]{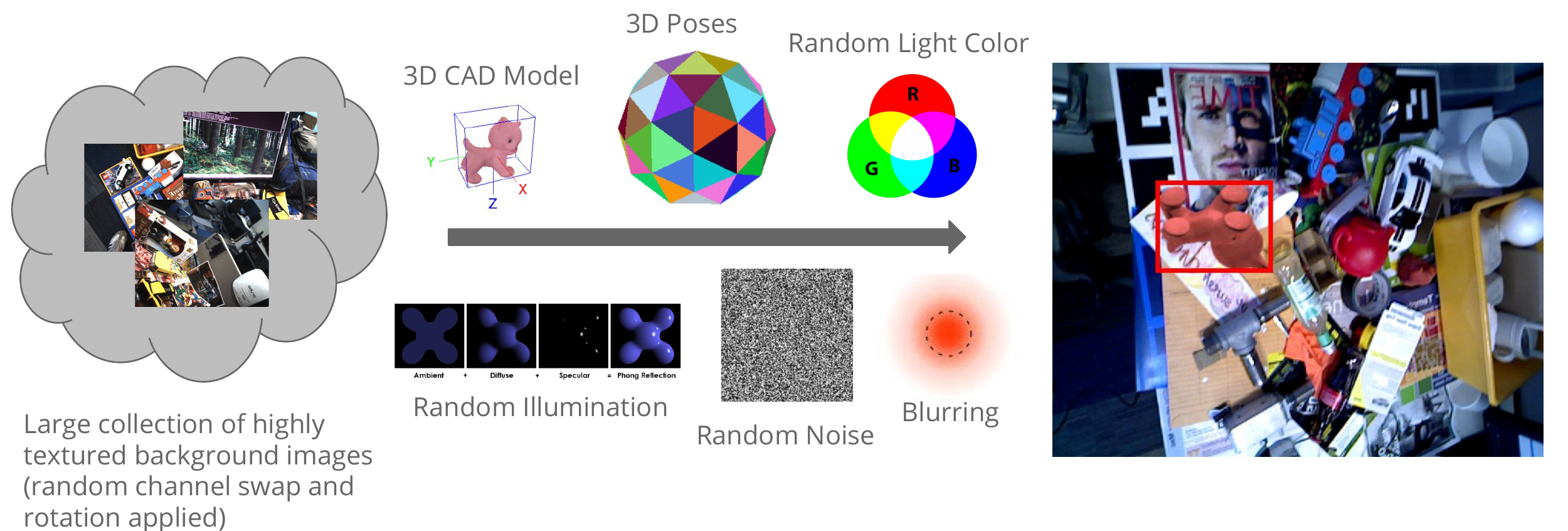} 
\end{center}
\caption{\label{fig:pipeline} Our synthetic data  generation pipeline.  For each
  generated 3D  pose and object, we  render the object over  a randomly selected
  cluttered   background  image   using  OpenGL   and  the   Phong  illumination
  model~\cite{Phong75}.  We use randomly perturbed light color for rendering and
  add image noise to the rendering.  Finally, we blur the object with a Gaussian
  filter. We also compute a tightly  fitting bounding box using the object's CAD
  model and the corresponding pose.}

\end{figure*}


Similar to~\cite{Dwibedi17},  we believe  that while  global consistency  can be
important, local appearance---so called patch-level realism---is also important.
The term patch-level  realism refers to the observation that  the content of the
bounding box framing the rendered object  looks realistic.

This  principle is  an important  assumption for  our synthetic  data generation
pipeline,  shown  in Fig.~\ref{fig:pipeline}.   For  each  object, we  start  by
generating a large  set of poses uniformly  covering the pose space  in which we
want    to    be   able    to    detect    the   corresponding    object.     As
in~\cite{Hinterstoisser12b}, we  generate rotations  by recursively  dividing an
icosahedron, the largest convex regular polyhedron.  We substitute each triangle
into four almost equilateral triangles, and iterate several times.  The vertices
of the  resulting polyhedron give us  then the two out-of-plane  rotation angles
for the  sampled pose  with respect  to the coordinate  center.  In  addition to
these  two  out-of-plane  rotations,  we   also  use  equally  sampled  in-plane
rotations.
Furthermore, we  sample the  scale logarithmically  to guarantee  an approximate
linear change  in pixel coverage  of the reprojected object  between consecutive
scale levels. 

The object is rendered at a random location in a randomly selected
background image using a uniform  distribution. The selected background image is
part of  a large collection  of highly cluttered real  background images taken
with the camera  of choice where the objects  of interest are
not included.
To increase  the variability of the  background image set, we  randomly swap the
three   background   image  channels   and   randomly   flip  and   rotate   the
images~($0\degree$, $90\degree$,  $180\degree$ and $270\degree$).  
We also tried to work without using real background images 
and experimented with backgrounds only exhibiting one randomly chosen color, however,
that didn't lead to good results.

We  use plain OpenGL with simple Phong  shading~\cite{Phong75}   for  rendering  where  we   allow  small  random
perturbations of the ambient, the diffuse  and the specular parameters.  We also
allow small  random perturbations  of the  light color.   We add  random Gaussian
noise to the rendered  object and blur it with a  Gaussian kernel, including its
boundaries  with  the  adjacent  background  image pixels  to  better
integrate  the  rendering  with  the  background.   We  also  experimented  with
different   strategies   for  integrating   the   rendered   object  in   images
as~\cite{Dwibedi17},  however this  did  not result  in significant  performance
improvements.




\subsection{Freezing a Pre-Trained Feature Extractor}
\label{sec:enabling_synthetic_data}

As shown  in~\cite{Huang17} and illustrated  in Fig.~\ref{fig:metaarchitecture},
many    state-of-the-art    object     detectors    including    
Faster-RCNN~\cite{faster_rcnn},          Mask-RCNN~\cite{mask_rcnn},         and
R-FCN~\cite{rfcn}  can  be decoupled  as  a  'meta-architecture' and  a  feature
extractor     such     as    VGG~\cite{vgg},     Resnet~\cite{resnet101},     or
InceptionResnet~\cite{inception_resnet}.

While  the meta-architecture  defines the  different modules  and how  they work
together,  the  feature  extractor  is  a deep  network  cut  at  some  selected
intermediate convolutional level. The remaining part  can be used as part of the
multi-way classification+localization  of the object detector.   As discussed in
the introduction, for the feature extractor,  we use frozen weights pre-learned on real
images, to  enable training  the remaining part  of the  architecture on  synthetic images
only.



In    practice,   we    use   the    Google's   public    available   OpenSource
version~\cite{Huang17} of  Faster-RCNN and RFCN,  and our own  implementation of
Mask-RCNN. The 'frozen' parts are taken according to~\cite{Huang17}, by training
InceptionResnet  and Resnet101  on  a classification  task  on the  ImageNet-CLs
dataset.
\comment{
\vincentrmk{Can you give more details  on this classification task? Maybe
  a  reference?} The reference is the same as Huang17 - I asked them about details and this is what I got back:
In some cases (VGG, Resnet), where the network originated from outside Google, we convert Caffe weights from the original authors.  In cases where the network was trained within Google like Inception or Mobilenet, then it's always through slim and is equivalent to the imagenet training code that we've released to open source.
}  
We  freeze  InceptionResnet~(v2) after  the  repeated use  of
block17  and right  before layer  Mixed$\_$7a and  Resnet101 after  block3.  All
other remaining  parts of the networks  are not 'frozen', meaning  their weights
are free to adapt when we train the detector on synthetic images.

We evaluate this approach in the next section.

\begin{figure}
\begin{center}
\begin{tabular}{cc}
\includegraphics[height=0.42\linewidth]{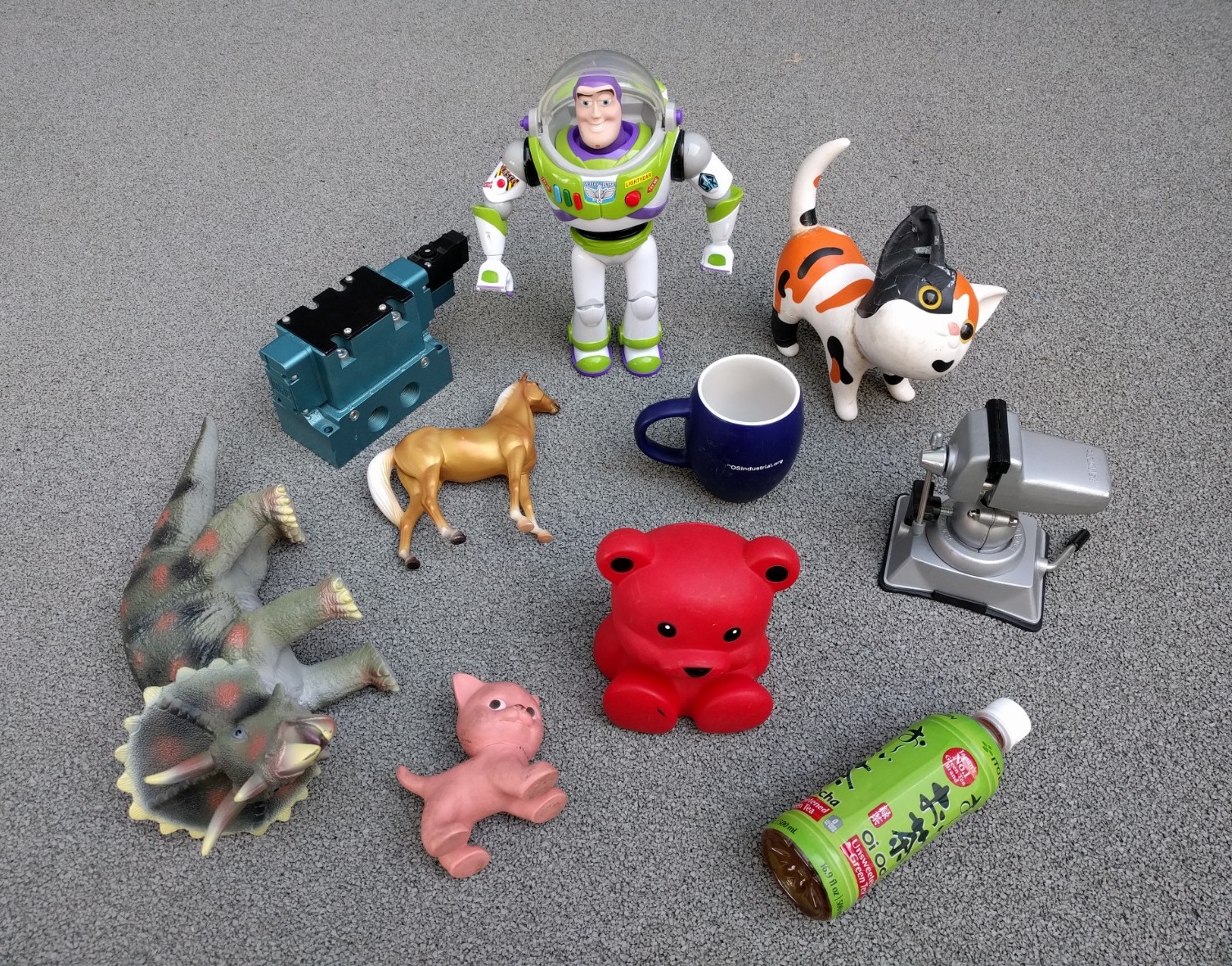} &
\includegraphics[height=0.42\linewidth]{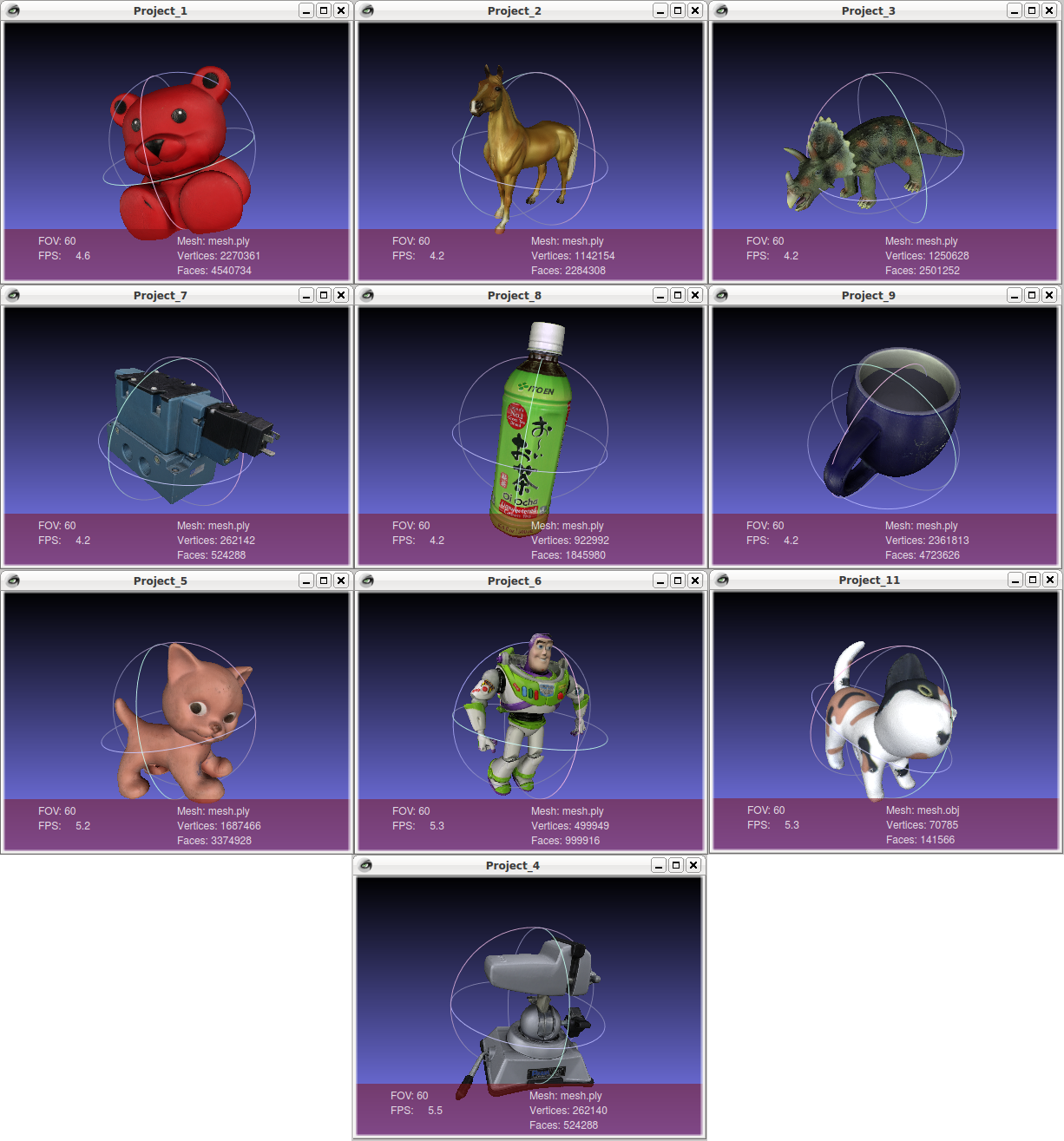} \\
(a) Real Objects &
(b) 3D CAD Models \\
\end{tabular}
\end{center}
\caption{\label{fig:models} (a) The real objects used in our experiments and
(b)  their CAD  models.  We chose  our  objects
  carefully to represent  different colors and 3D shapes and  to cover different
  fields of applications (industrial objects, household objects, toys). }
\end{figure}

\section{Experiments}
\label{sec:epxeriments}

In this section, we first describe the dataset we created for these evaluations,
made of synthetic  and real images of 10 different  objects.  We also considered
two different cameras,  as the quality of the camera  influences the recognition
results as we  will show.  The rest  of the section reports  our experiments and
the conclusions we draw from them.

\subsection{Objects and 3D CAD models}
\label{lab:objects}

As shown in Fig.~\ref{fig:models}, we carefully  selected the objects we used in
our experiments: We  tried to represent different  colors, textures (homogeneous
color versus  highly textured),  3D shapes  and material  properties (reflective
versus non-reflective).  Except for the mug and the bottle, the 3D shapes of the
objects we selected can look very different from different views.  We also tried
to  consider  objects from  different  application  fields (industrial  objects,
household objects, toys).  For each real object  we have a textured 3D CAD model
at hand which we generated using our in-house 3D scanner.

\subsection{Cameras}
We consider two cameras, an \asus and a \ptgrey.  For each camera, we generated
a training dataset and an evaluation  dataset.  The training datasets consist of
approximatively 20K and  the evaluation datasets of  approximatively 1K manually
labeled real  world images.  Each  sample image contains  one of the  10 objects
shown  in Fig.~\ref{fig:models}  in challenging  environments: heavy  background
clutter, illumination changes, etc.  In addition,  we made sure that each object
is  shown from  various poses  as  this is  very important  for object  instance
detection.  Furthermore,  for each dataset all  objects have the same  amount of
images.    

\begin{figure}
\begin{center}
\begin{tabular}{c}
\includegraphics[width=0.9\linewidth]{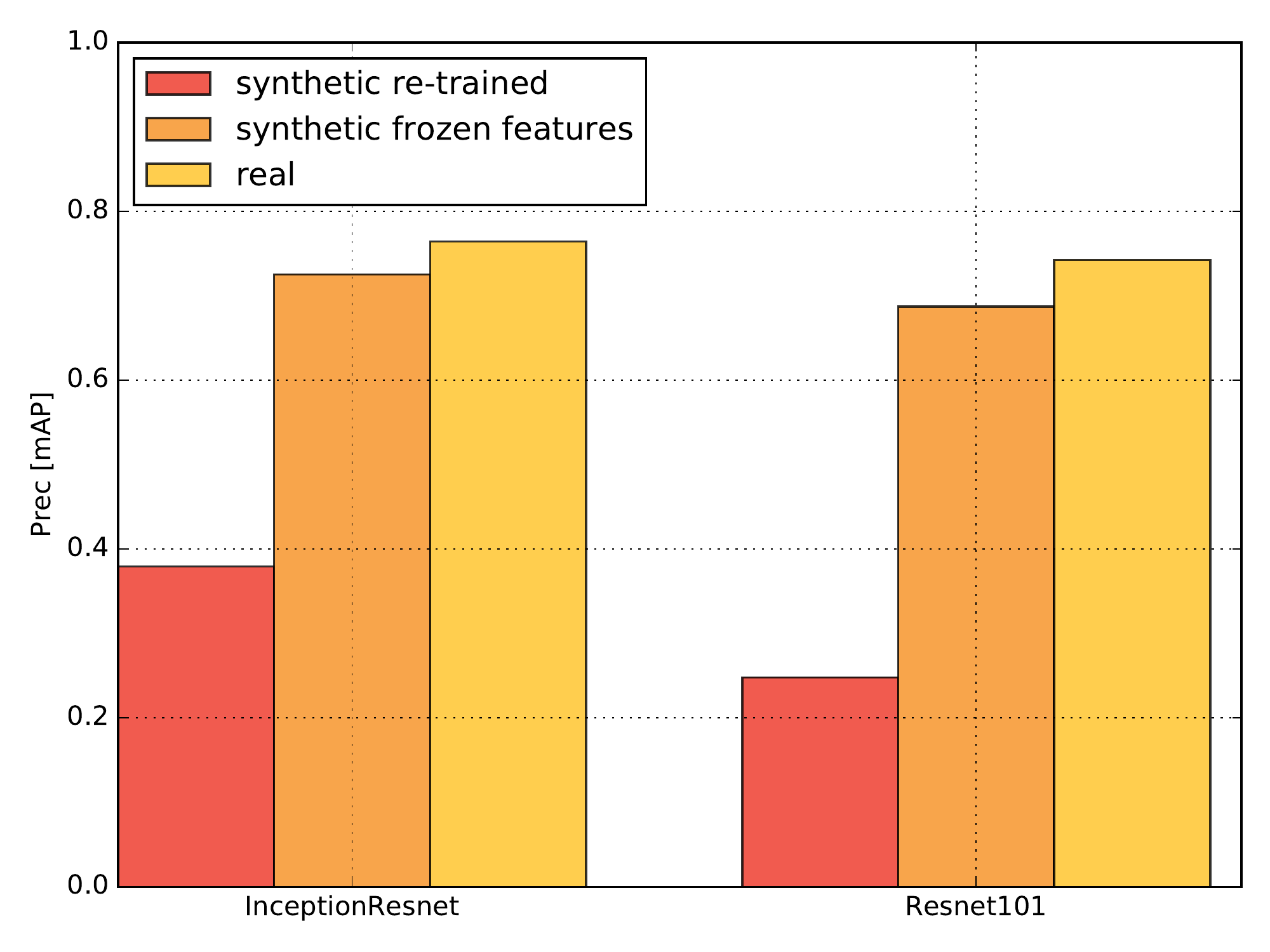} \\
(a) \ptgrey \\
\includegraphics[width=0.9\linewidth]{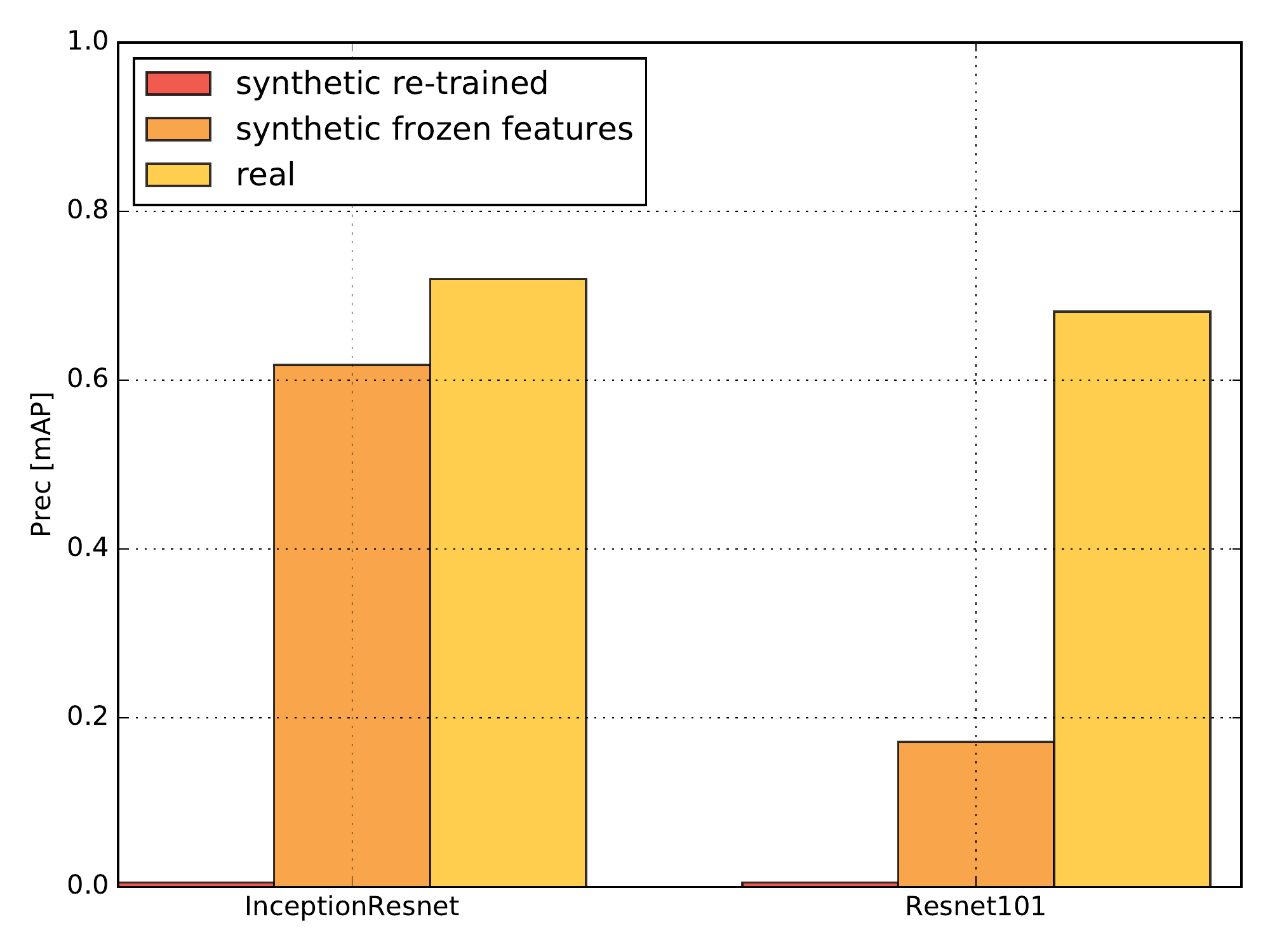} \\
(b) \asus \\
\end{tabular}
\end{center}
\caption{\label{fig:freezing}  The effect  of freezing  the pre-trained  feature
  extractor,  for two  different cameras.   Training the  feature extractors  on
  synthetic images  performs poorly, and totally  fails in the case  of the \asus
  camera.  When  using  feature  extractors   pre-trained  on  real  images  without
  retraining them, the performances of detectors trained on synthetic data are almost  as good as when training them on real
  data, except when ResNet101 is used with images from the \asus camera.}
\end{figure}

\subsection{Freezing the Feature Extractor}
\label{subsec:freezing}

Fig.~\ref{fig:freezing}  shows that  when  Faster-RCNN is  trained on  synthetic
images and  tested on  real images,  it performs  significantly worse  than when
trained  on real  data.  By contrast,  when we  freeze  the feature  extractor's
weights during training to values pre-trained on real images, and only train the
remaining parts  of the detector,  we get  a significant performance  boost.  We
even come close to detectors trained purely  on real world data, as we typically
obtained up to $95\%$ of the performance when trained on synthetic data.



\begin{figure}
\begin{center}
\begin{tabular}{@{\hspace{-0.2cm}}c@{}cc@{}c}
\includegraphics[width=0.25\linewidth]{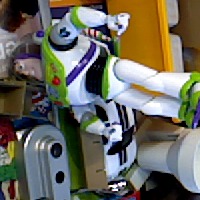} &
\includegraphics[width=0.25\linewidth]{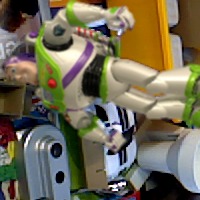} &
\includegraphics[width=0.25\linewidth]{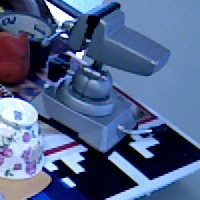} &
\includegraphics[width=0.25\linewidth]{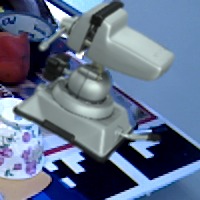} \\
\end{tabular}
\begin{tabular}{c}
\includegraphics[width=0.8\linewidth]{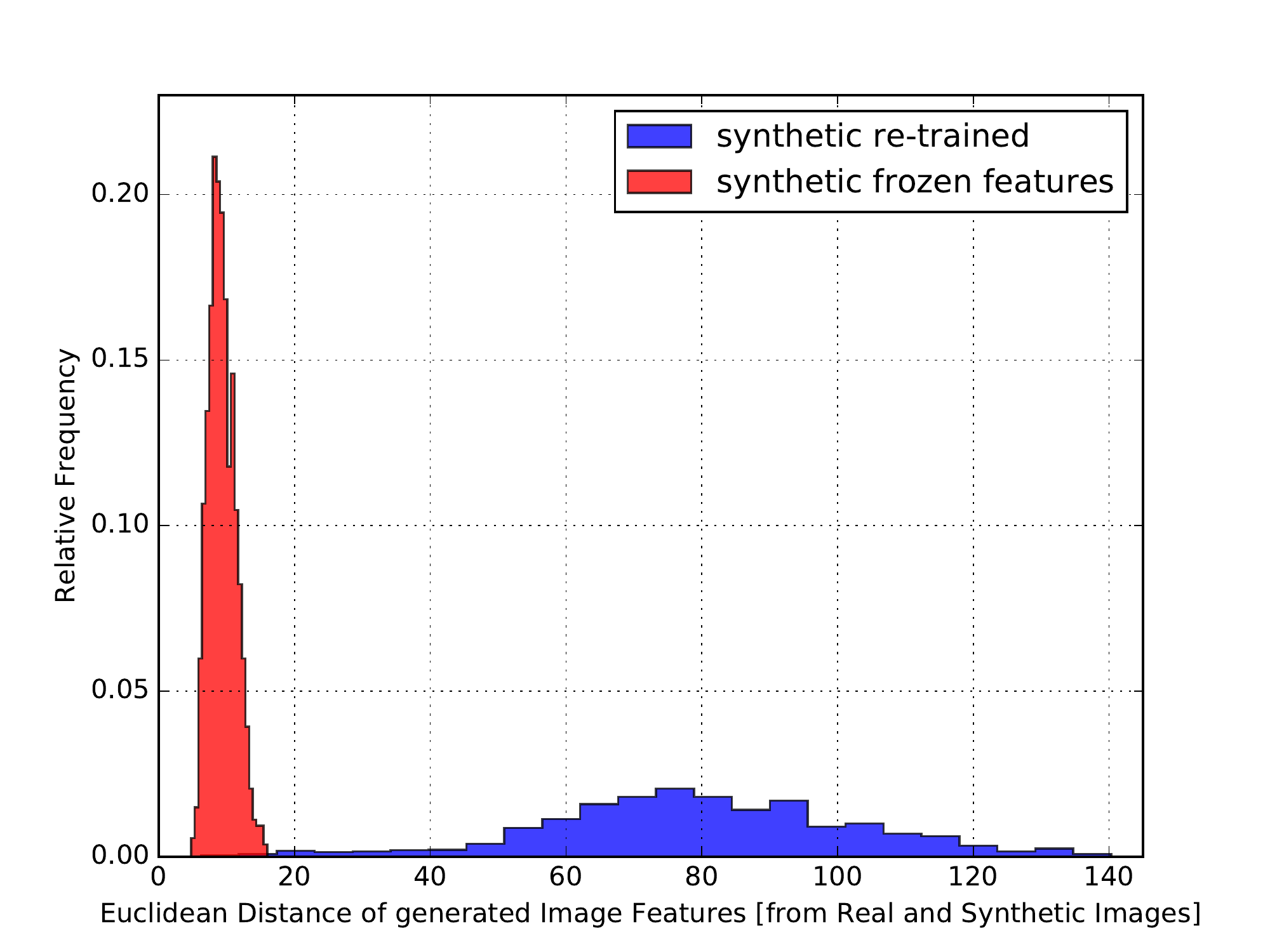} 
\end{tabular}
\end{center}
\caption{\label{fig:euclidean} Top: Two examples of pairs  of a real image and a
  synthetic one for  the same object under the same  pose. Bottom: Distributions
  of  the Euclidean  distances between  image  features generated  for the  real
  images and the corresponding synthetic images. See text in
  Section~\ref{subsec:freezing} for details.}
\end{figure}

To get a better intuition why  freezing the feature extractor gives significant
better results than retraining it on  synthetic data, we performed the following
experiment: We  created 1000  image pairs with  different objects  under various
poses.  Each image pair consists of one  image that shows the real object and of
another image where we superpose a rendering of the object's CAD model on top of
the real image, under the same pose as the real object. The top of
Fig.~\ref{fig:euclidean} shows two examples. 

We  then compared  the distributions  of the  Euclidean distances  between image
features generated for  the real images and the  corresponding synthetic images.
As we  can see at  the bottom  of Fig.~\ref{fig:euclidean}, the  distribution is
much more  clustered around  0 when  the features are  computed using  a frozen feature
extractor pre-trained  on real  images (red) compared  to the  distribution obtained
when the pre-trained feature extractor is finetuned on synthetic images (blue).


\label{sec:layers}
\begin{figure*}[ht]
\begin{center}
\begin{tabular}{cc}
\includegraphics[width=0.4\linewidth]{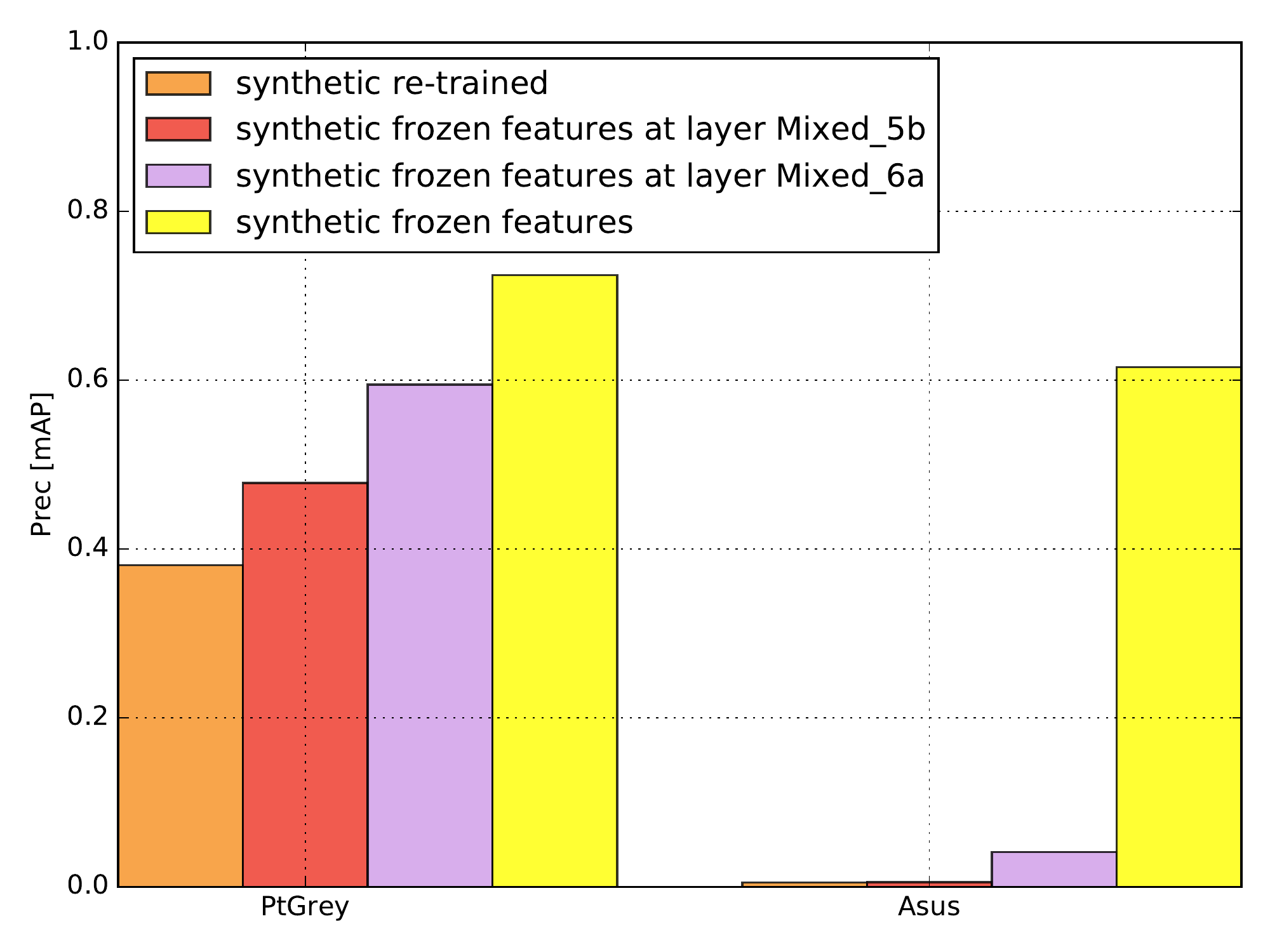} &
\includegraphics[width=0.4\linewidth]{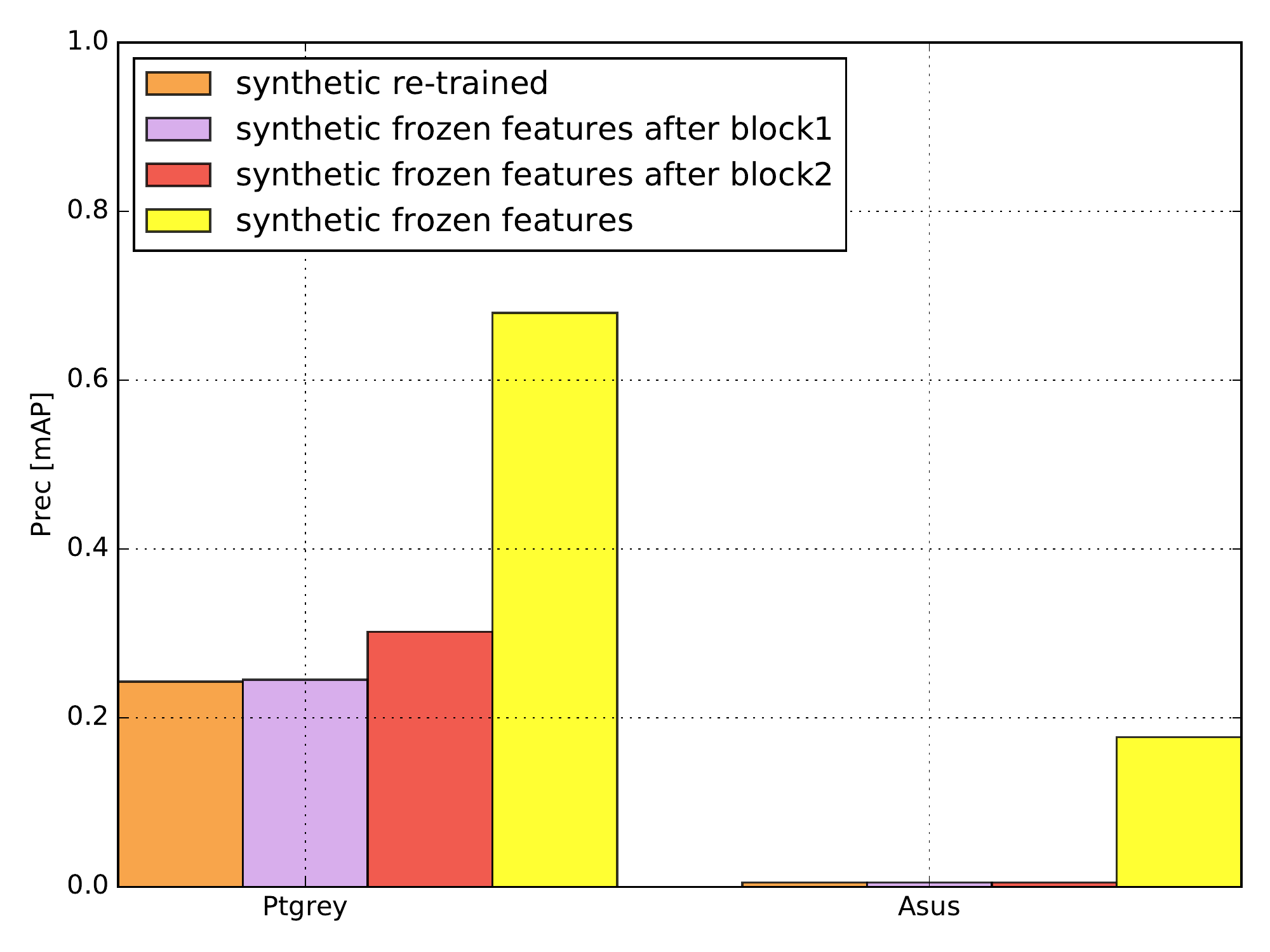} \\
a) InceptionResnet~\cite{inception_resnet} &
b) Resnet101~\cite{resnet101} \\
\end{tabular}
\end{center}
\caption{\label{fig:layers}   We  freeze   features  at   different  layers   of
  InceptionResnet~\cite{inception_resnet}  and  Resnet101~\cite{resnet101}.   We
  can see that freezing the full feature extractor performs best (yellow).}
\end{figure*}

\subsection{Freezing the Feature Extractor at Different Layers}
We also performed experiments where we freeze the feature extractor at different
intermediate layers i.e. layers lying between the input and the output layers of
the feature extractor as specified in Section~\ref{sec:enabling_synthetic_data}.
As can  be seen  in Fig.~\ref{fig:layers}, freezing  the full  feature extractor
always performs best.   For the \asus camera, freezing the  feature extractor on
intermediate levels even results in a dramatic loss of performance.



\subsection{On Finetuning the Feature Extractor}
\label{sec:unfreezing}

\begin{figure*}[ht]
\begin{center}
\begin{tabular}{cc}
\includegraphics[width=0.4\linewidth]{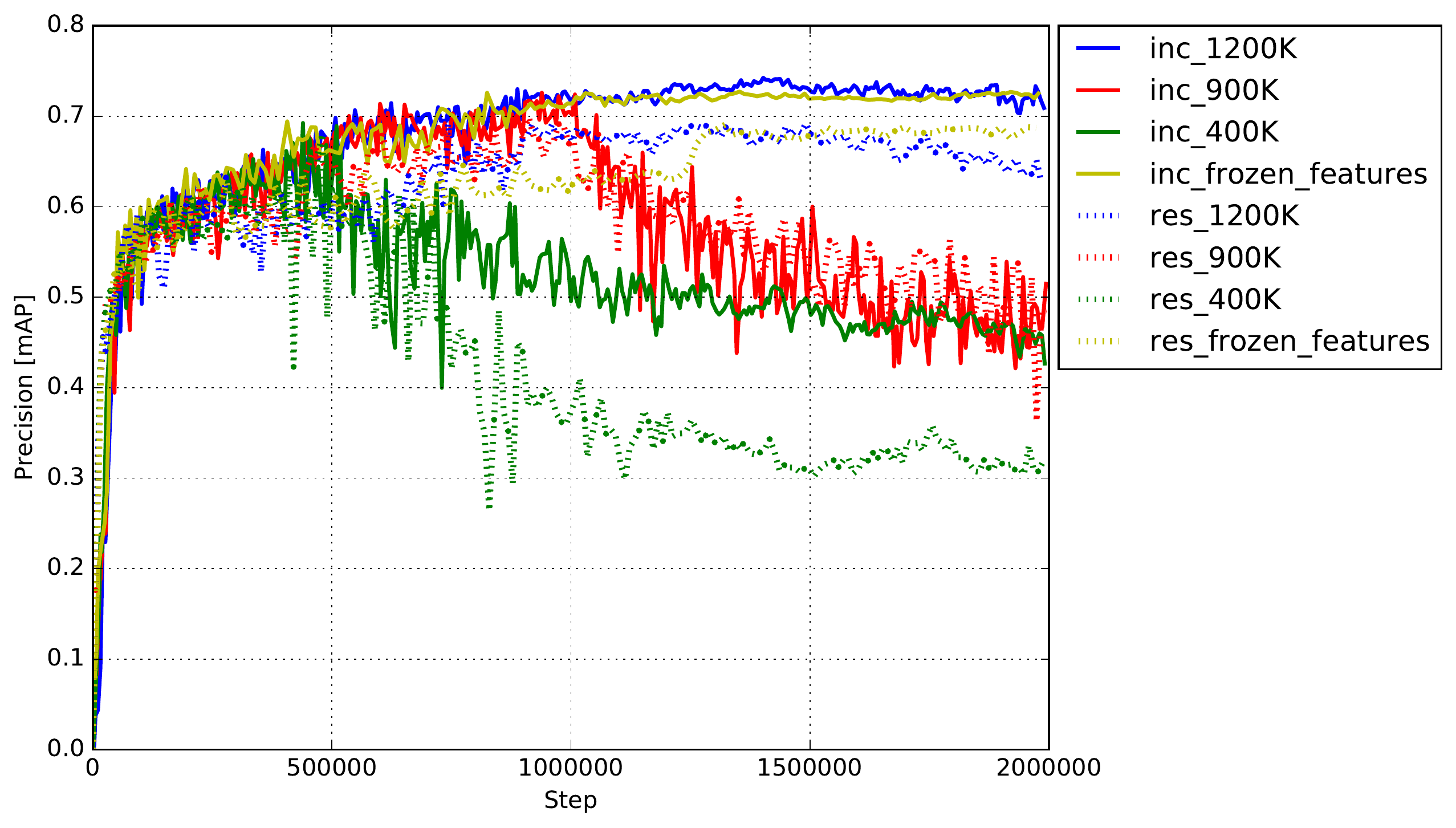} &
\includegraphics[width=0.4\linewidth]{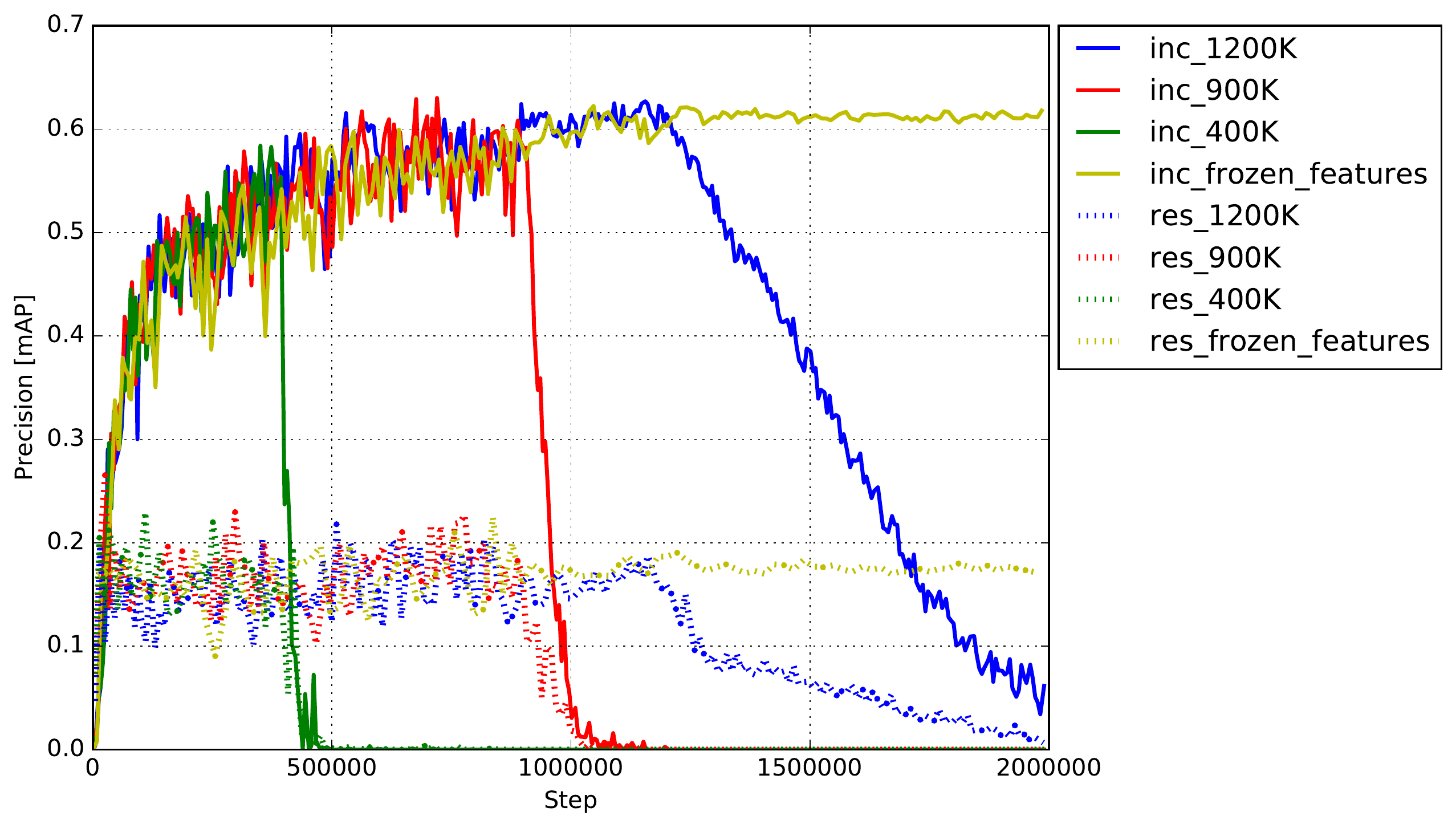} \\
a) \ptgrey &
b) \asus \\
\end{tabular}
\end{center}
\caption{\label{fig:incremental}  Finetuning the  feature extractor  after 400K,
  900K and 1200K steps where the pre-trained feature extractor was frozen for the \ptgrey and the \asus  cameras.  We show results
  for the InceptionResnet~\cite{inception_resnet} and Resnet101~\cite{resnet101}
  architectures.}
\end{figure*}

One may wonder if the domain shift between synthetic and real images still leads
to decreased performance  after the detector was trained for  some time with the
pre-trained feature extractor frozen.  One could argue that all remaining detector weights
have already started to converge and therefore, the domain shift \comment{does not matter
any more} is far less influential.  As a  result, the frozen feature extractor could  be unfrozen 
to finetune its weights to adapt to the learning task.

However, as we show in Fig.~\ref{fig:incremental},  this is not true.  Even 
after  1200K training  steps  where the  feature extractor  was  frozen and  the
detection  performance starts  to  plateau the  detector's performance  degrades
significantly    if    the    frozen     feature    extractor    is    unfrozen and its weights are finetuned.
Table~\ref{tab:main_exp} gives the corresponding numbers.

\begin{table*}
\begin{center}
\small {
\begin{tabular}{@{}llccccc|c@{}}
\toprule
\multicolumn{1}{c}{} & \multicolumn{1}{c}{} & \multicolumn{1}{c}{synthetic} & \multicolumn{1}{c}{frozen} & \multicolumn{1}{c}{400K} & \multicolumn{1}{c}{900K} & \multicolumn{1}{c}{1200K} & \multicolumn{1}{c}{real} \\
\midrule
\multirow{4}{*}{\rotatebox[origin=c]{90}{Asus}}
& Prec [mAP]     &  \blue{.000}/\red{.000} & \blue{\bf{.617}}/\red{.171} & \blue{.000}/\red{.000} & \blue{.000}/\red{.000} & \blue{.061}/\red{.006} & \blue{.719}/\red{.681} \\
& Prec [mAP@0.5] &  \blue{.000}/\red{.000} & \blue{\bf{.948}}/\red{.385} & \blue{.000}/\red{.000} & \blue{.000}/\red{.000} & \blue{.114}/\red{.016} & \blue{.983}/\red{.988} \\
& Prec [mAP@0.75] & \blue{.000}/\red{.000} & \blue{\bf{.733}}/\red{.130} & \blue{.000}/\red{.000} & \blue{.000}/\red{.000} & \blue{.064}/\red{.004} & \blue{.872}/\red{.844} \\
& Acc [@100] &      \blue{.000}/\red{.010} & \blue{\bf{.686}}/\red{.256} & \blue{.000}/\red{.000} & \blue{.000}/\red{.000} & \blue{.079}/\red{.007} & \blue{.772}/\red{.742} \\
\midrule
\multirow{4}{*}{\rotatebox[origin=c]{90}{PtGrey}} 
& Prec [mAP]     &  \blue{.374}/\red{.243} & \blue{\bf{.725}}/\red{.687} & \blue{.426}/\red{.317} & \blue{.514}/\red{.485} & \blue{.709}/\red{.626} & \blue{.764}/\red{.742} \\
& Prec [mAP@0.5] &  \blue{.537}/\red{.410} & \blue{\bf{.971}}/\red{.966} & \blue{.606}/\red{.491} & \blue{.717}/\red{.685} & \blue{.936}/\red{.912} & \blue{.987}/\red{.987} \\
& Prec [mAP@0.75] & \blue{.431}/\red{.239} & \blue{\bf{.886}}/\red{.844} & \blue{.495}/\red{.355} & \blue{.593}/\red{.564} & \blue{.835}/\red{.756} & \blue{.908}/\red{.916} \\
& Acc [@100] &      \blue{.461}/\red{.324} & \blue{\bf{.771}}/\red{.736} & \blue{.483}/\red{.384} & \blue{.577}/\red{.551} & \blue{.768}/\red{.695} & \blue{.808}/\red{.804} \\
\bottomrule
\end{tabular}
}
\end{center}
\caption{\label{tab:main_exp} Outcomes  of all our experiments.  We give numbers
  for              \blue{InceptionResnet}~\cite{inception_resnet}              /
  \red{Resnet101}~\cite{resnet101}.  Except  for the experiments with  real data
  (last column),  all experiments  were performed   on synthetic  data only. We
  emphasized the best results trained on synthetic data.}
\end{table*}

\subsection{Ablation Experiments}
\label{sec:influence}

\begin{figure*}[ht]
\begin{center}
\begin{tabular}{cc}
\includegraphics[width=0.4\linewidth]{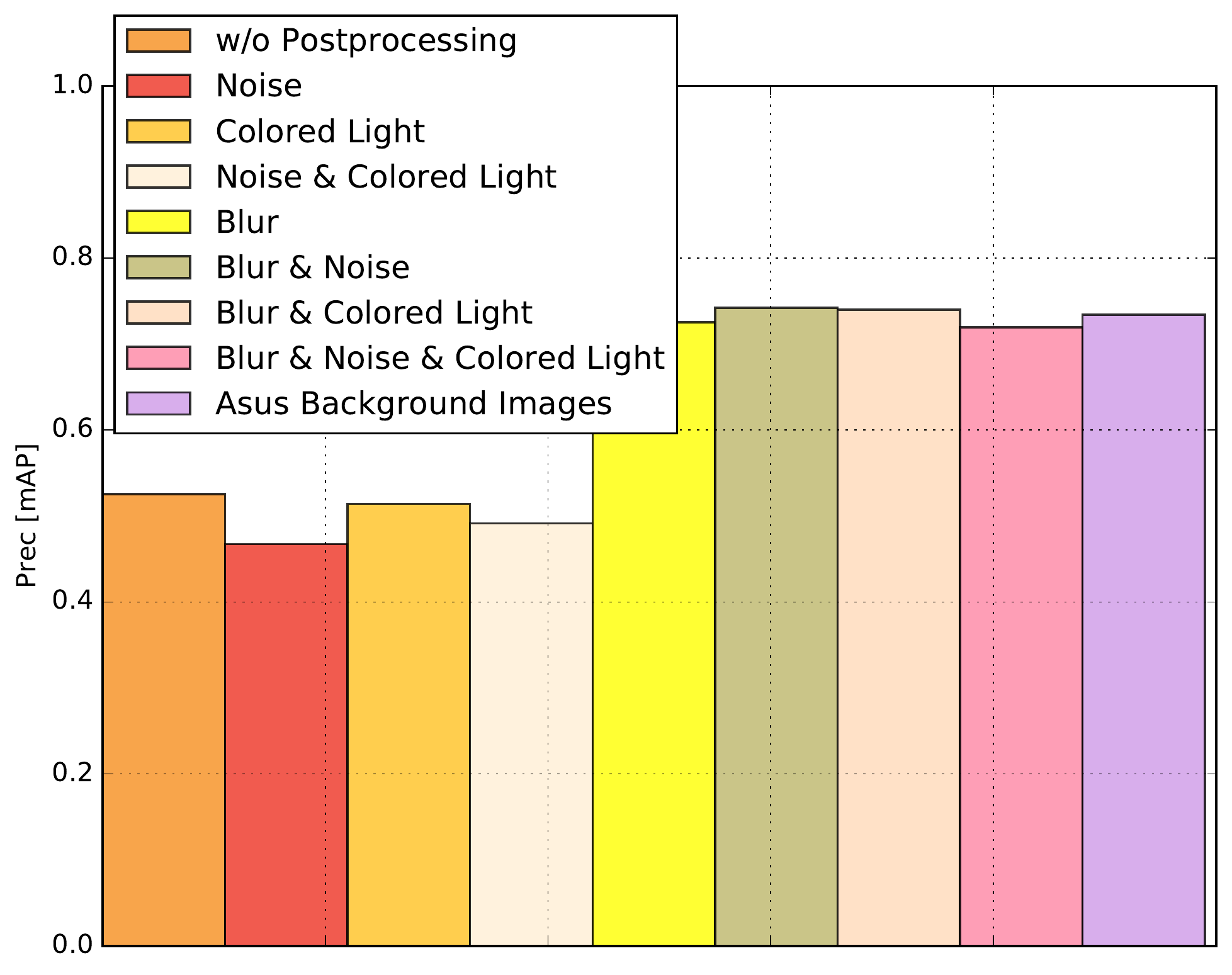} &
\includegraphics[width=0.4\linewidth]{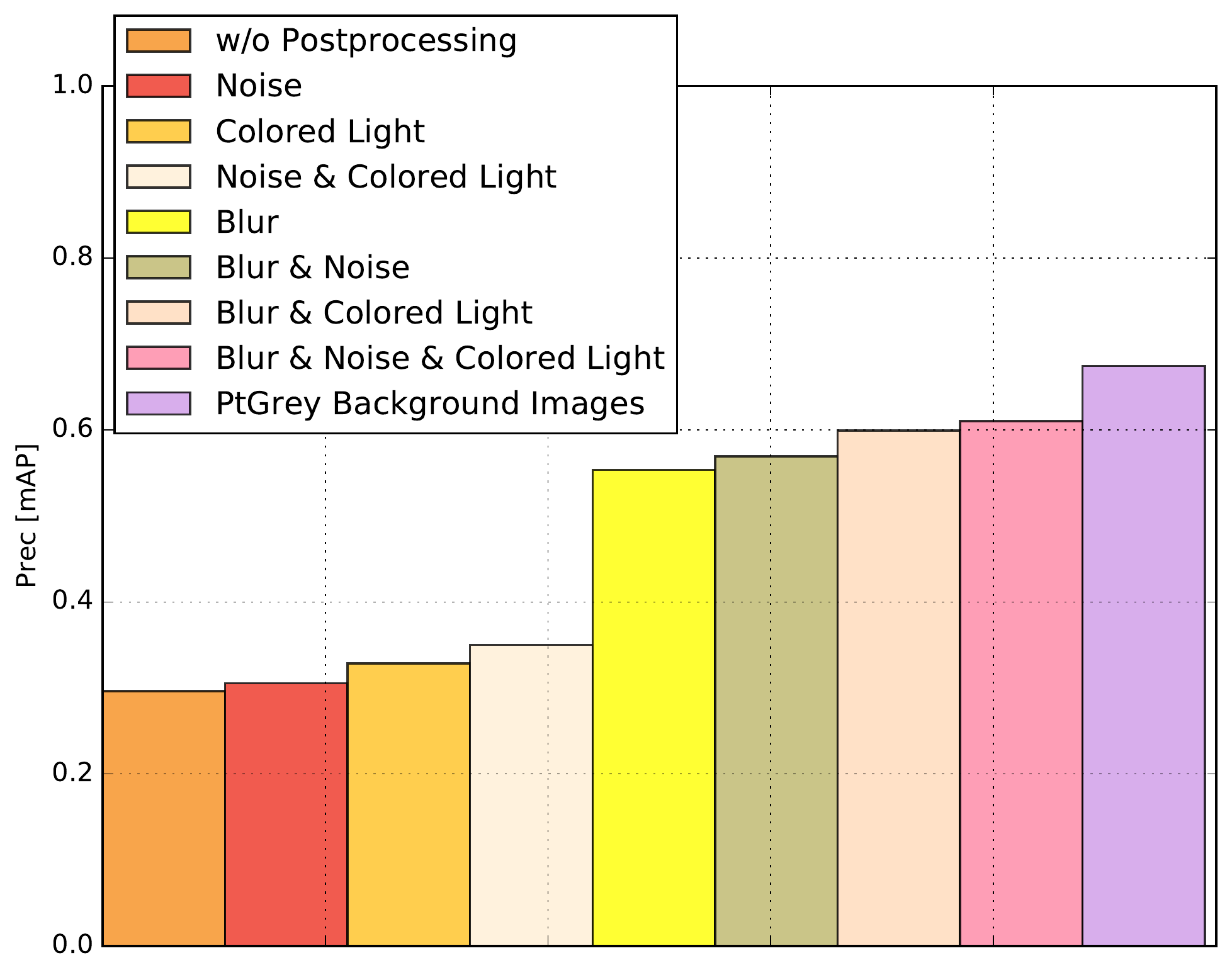} \\
a) \ptgrey &
b) \asus \\
\end{tabular}
\end{center}
\caption{\label{fig:influence} Influences  of the different building  blocks for
  synthetic  rendering for  the \ptgrey  and  the \asus  cameras.  Results  were
  obtained   with   InceptionResnet~\cite{inception_resnet}   as   the   feature
  extractor. Blurring is clearly a useful yet simple operation to apply to the
  synthetic images to improve the results.}
\end{figure*}

In the following experiments, we investigated  the influence of the single steps
in  the  image   generation  pipeline.   For  all  these   experiments  we  used
InceptionResnet~\cite{inception_resnet}  as  feature   extractor.   The  feature
extractor itself was frozen.  We found out that blurring the rendered object and
its adjacent image  pixels gives a huge performance boost.   Adding noise to the
rendered object or enabling random light  color did not give much improvement in
performance and its influence depends on the camera used.  As already mentioned,
we also experimented with  different blending strategies as in~\cite{Dwibedi17},
that  is using  different blending  options in  the same  dataset: no  blending,
Gaussian blurring  and Poisson blending,  however we could not  find significant
performance improvements.

We also investigated what happens if we use the internal camera parameter of our
target camera but a background dataset taken with another camera.  As we can see
in Fig.~\ref{fig:influence} results seem to  stay approximately the same for the
\ptgrey  camera and  seem  to improve  for  the \asus  camera.  The later  seems
reasonable since  the background images taken  with the \ptgrey camera  are more
cluttered and  are showing  more background variety  than the  background images
taken with the  \asus camera.  These results suggest that  the camera images can
be taken  from an arbitrary  source and we  only have to  make sure that  a high
amount of background variety is provided.


\subsection{RFCN and MASK-RCNN}
\label{sec:other_detectors}

\begin{figure}[ht]
\begin{center}
\includegraphics[width=0.9\linewidth]{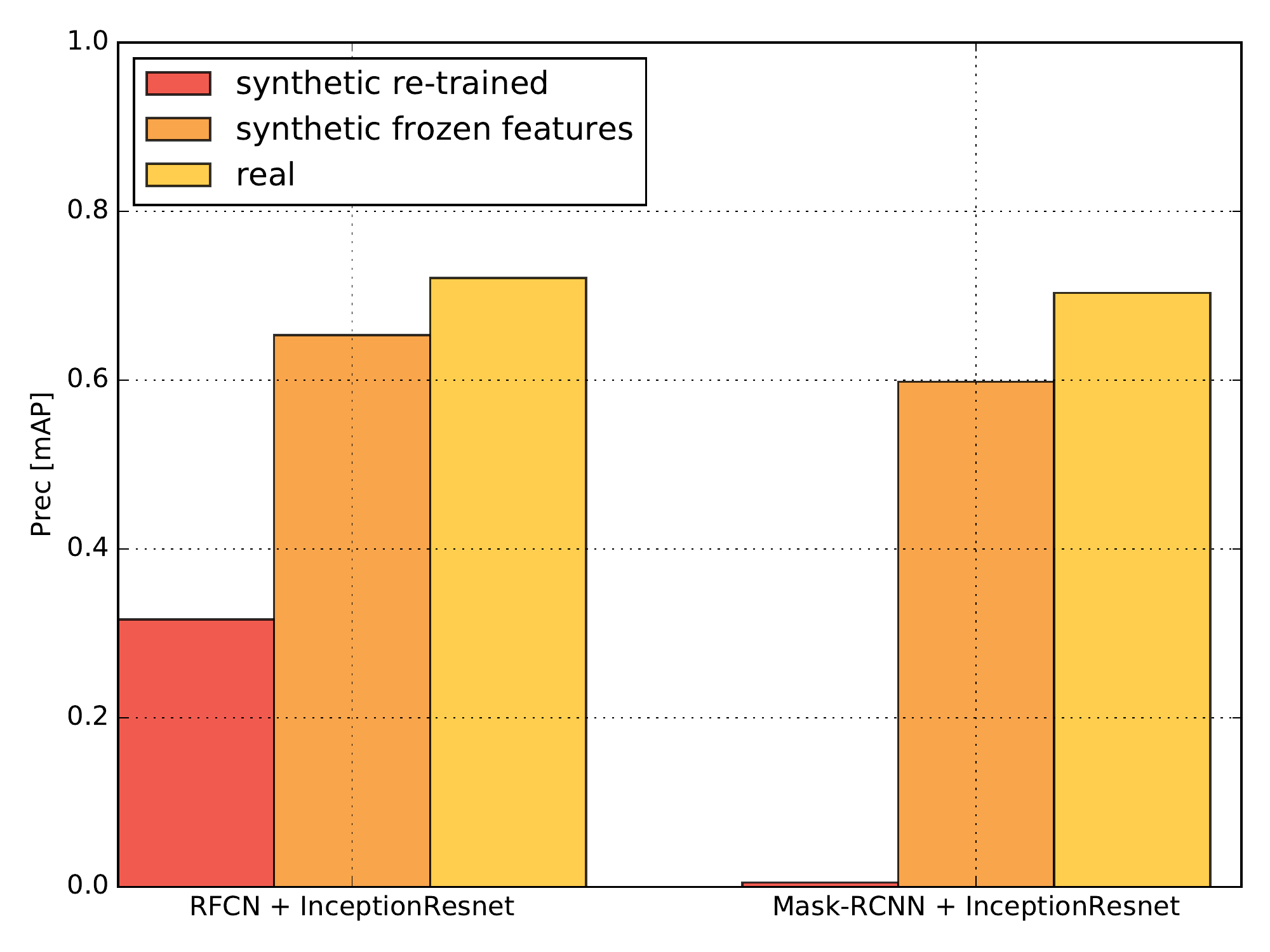}
\end{center}
\caption{\label{fig:rfcn}  Results   using  RFCN~\cite{rfcn}  on   the  \ptgrey
  dataset. Freezing  the feature  extractor boosts performance  significantly on
  this method as well.  We observe the same results if  we train Mask-RCNN on
    the \asus dataset.}
\end{figure}

To  show the  generality of  our approach,  we also  performed several  addition
experiments.  Fig.~\ref{fig:rfcn} shows the results for RFCN~\cite{rfcn} trained
only on  synthetic data with  the feature  extractor frozen and  compares them
with those  using RFCN  trained on  real data and  and those
using RFCN re-trained on synthetic data. Freezing the feature extractor helps to
unlock significant performance improvements also here.

\begin{figure*}
\begin{center}
\begin{tabular}{cccc}
\includegraphics[trim={0.55cm 0.5cm 0.6cm 1.35cm},clip=true,width=0.245\linewidth]{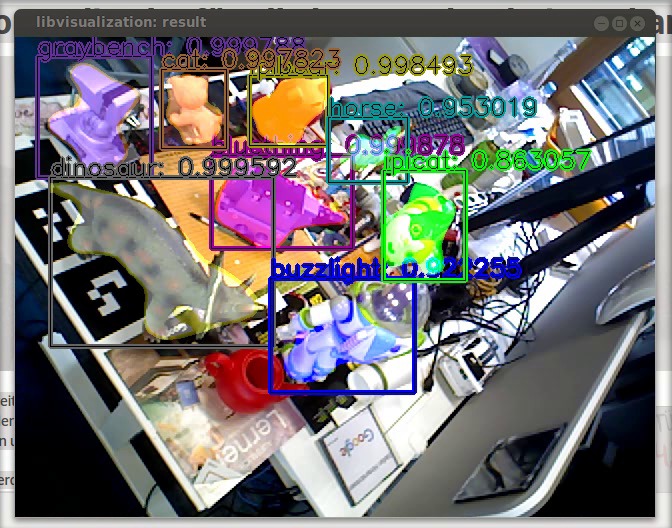} & \imgspace
\includegraphics[trim={0.55cm 0.5cm 0.6cm 1.35cm},clip=true,width=0.245\linewidth]{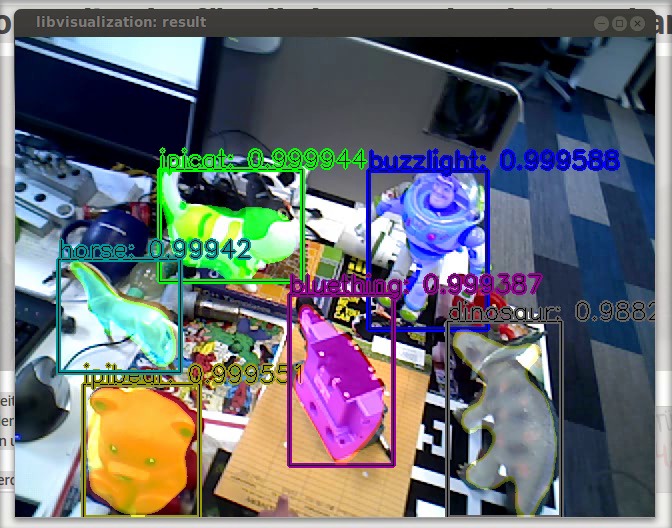} & \imgspace
\includegraphics[trim={0.55cm 0.5cm 0.6cm 1.35cm},clip=true,width=0.245\linewidth]{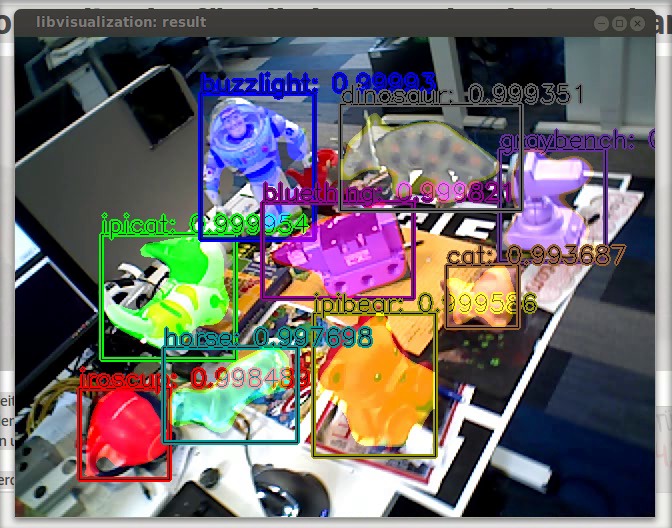} & \imgspace
\includegraphics[trim={0.55cm 0.5cm 0.6cm 1.35cm},clip=true,width=0.245\linewidth]{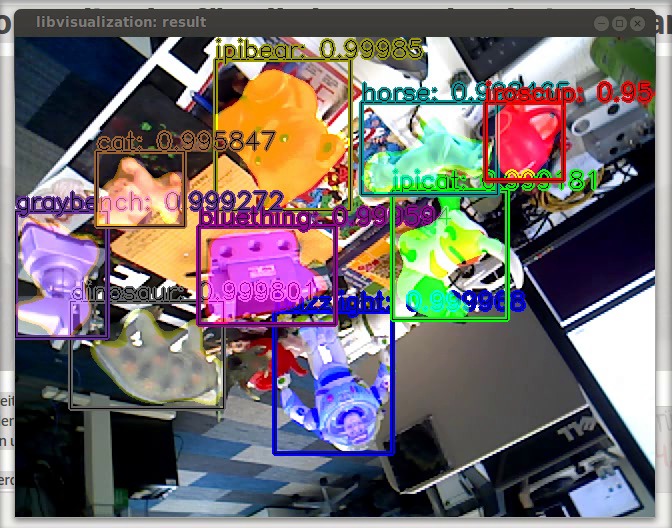} \\
\end{tabular}
\end{center}
\caption{\label{fig:mask_rcnn} Results of  Mask-RCNN~\cite{mask_rcnn} trained on
  synthetic images only with the feature extractor frozen. The images were taken
  with the \asus camera in a highly cluttered environment under various poses.}
\end{figure*}

Fig.~\ref{fig:rfcn} also shows quantitative  results of Mask-RCNN~\cite{mask_rcnn} trained only
on  synthetic data  with  the  feature extractor  frozen.   Similar  to what  we
observed with Faster-RCNN and RFCN, freezing the feature extractor significantly
boosts the performance when  trained on synthetic data. Fig.~\ref{fig:mask_rcnn}
shows that we are able to  detect objects in highly cluttered environments under
various poses  and get  reasonable masks.  This  result is  especially important
since it  shows that  exhaustive (manual)  pixel labeling  is made  redundant by
training from synthetic data.

\comment{Re-training Mask-RCNN  on  synthetic data results  in very  bad performance
similar as it is the case for Faster-RCNN and RFCN.}

\begin{figure*}
\begin{center}
\begin{tabular}{cccc}
\includegraphics[width=0.245\linewidth]{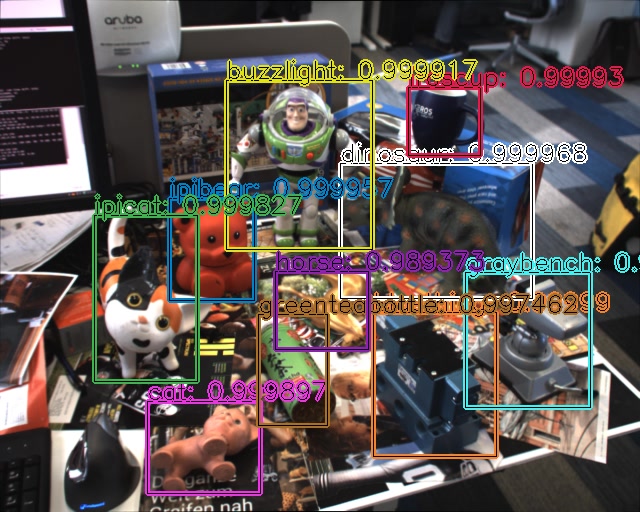} & \imgspace
\includegraphics[width=0.245\linewidth]{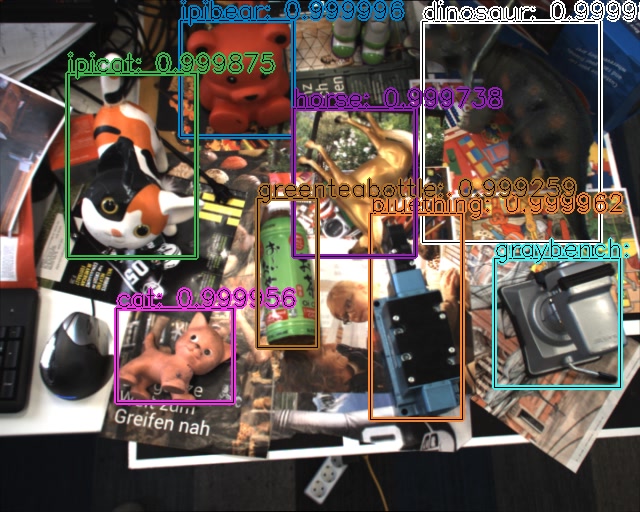} & \imgspace
\includegraphics[width=0.245\linewidth]{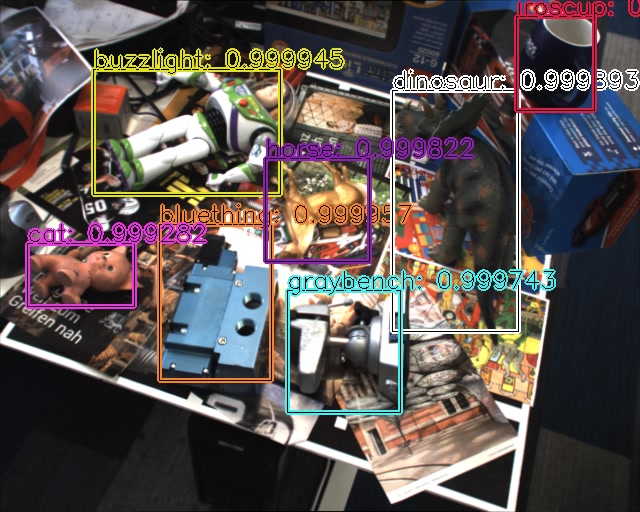} & \imgspace
\includegraphics[width=0.245\linewidth]{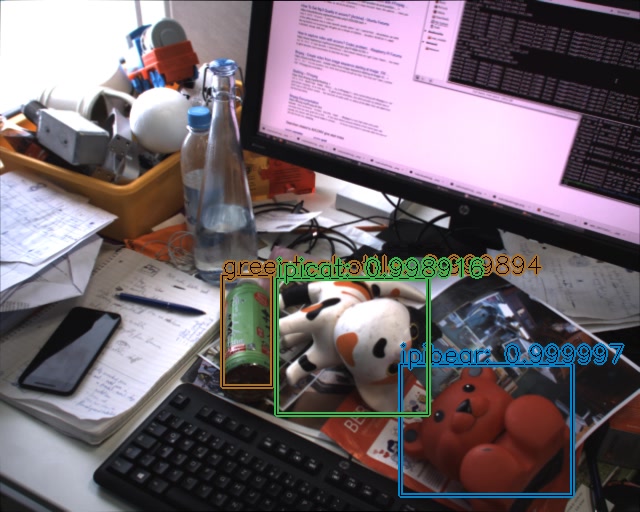} \\
\end{tabular}
\end{center}
\caption{\label{fig:results2} Results of Faster-RCNN trained on synthetic images
  only with  the feature extractor frozen.   The objects are detected  in highly
  cluttered  scenes and  many different  instances are  available in  one image.
  Note that the different objects are seen in different arbitrary poses.}
\end{figure*}

\begin{figure*}[!]
\begin{center}
\begin{tabular}{cccc}
\includegraphics[width=0.245\linewidth]{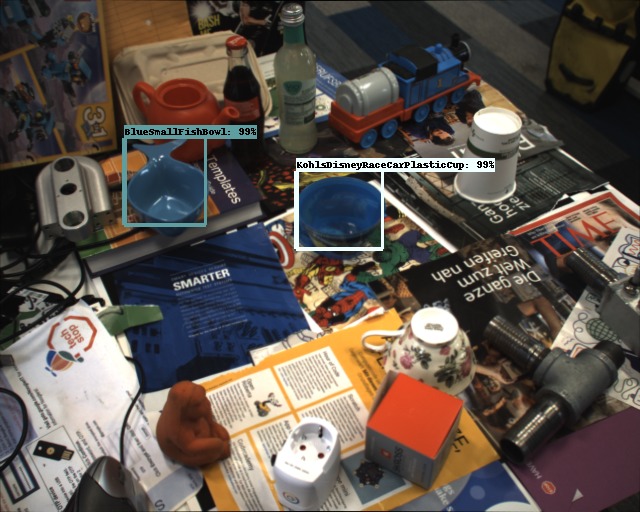} & \imgspace
\includegraphics[trim={0.55cm 1.0cm 0.6cm 2.7cm},clip=true,width=0.245\linewidth]{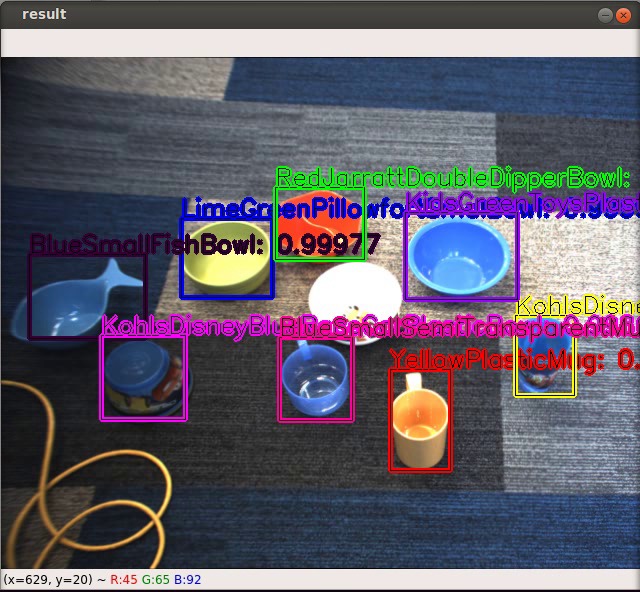} & \imgspace
\includegraphics[width=0.245\linewidth]{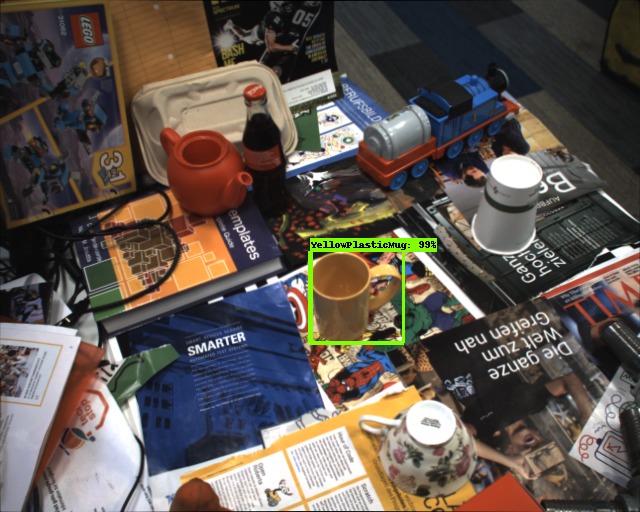} & \imgspace
\includegraphics[trim={0.55cm 1.0cm 0.6cm 2.7cm},clip=true,width=0.245\linewidth]{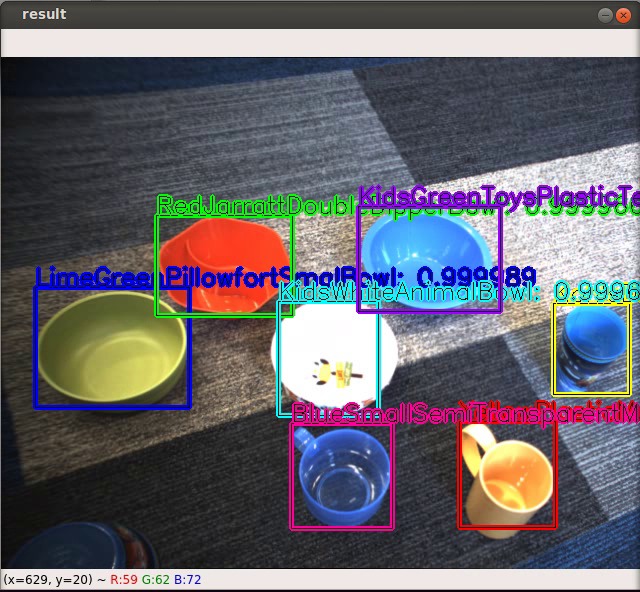} \\
\end{tabular}
\end{center}
\caption{\label{fig:other_objects} Objects with similar shapes and colors
  detected in challenging environments. The detector was trained on
  synthetic images only.}
\end{figure*}


\subsection{Qualitative Results}
\label{sec:images}

Fig.~\ref{fig:results2}  shows some  qualitative  results  on images  exhibiting
several of the  10 objects we considered in various  poses with heavy background
clutter         and         illumination        changes.          We         use Faster-RCNN~\cite{faster_rcnn} with the
InceptionResnet~\cite{inception_resnet}  as feature  extractor  and trained  the
rest of  the network on  synthetic images only.   Fig.~\ref{fig:mask_rcnn} shows
results    of     Mask-RCNN~\cite{mask_rcnn}    trained         on     synthetic
images only.
Fig.~\ref{fig:other_objects} shows  some other  objects trained with  the method
presented in this paper.

\section{Conclusion}
\label{sec:conclusions}

We have  shown that by  freezing a pre-trained feature  extractor we are  able to
train state-of-the-art object  detectors  on synthetic  data only.  The results
are close to  approaches trained on real data only.   While we have demonstrated
that  object  detectors   re-trained  on  synthetic  data   lead  to  poor
performances and that  images from different cameras lead  to different results,
freezing the feature extractor always gives a huge performance boost.

The results  of our experiments suggest  that simple rendering is  sufficient to
achieve good performances  and that complicated scene composition  does not seem
necessary.  Training  from rendered 3D  CAD models  allows us to  detect objects
from all possible viewpoints which makes the need for a real data generation and
expensive manual labeling pipeline redundant.

  
\textbf{Acknowledgments:} The authors thank Google's VALE team for tremendous support using the Google Object Detection API, especially Jonathan Huang, Alireza Fathi, Vivek Rathod, and Chen Sun. In addition, we thank Kevin Murphy, Vincent Vanhoucke, Konstantinos Bousmalis and Alexander Toshev for valuable discussions and feedback.

{
\small
\bibliographystyle{ieee}
\bibliography{./string,./cleaned_vision}
}

\end{document}